# On the visual analytic intelligence of neural networks

Stanisław Woźniak[1]*, Hlynur Jónsson[1,2]†, Giovanni Cherubini[1],
Angeliki Pantazi[1], Evangelos Eleftheriou[1]‡

[1]IBM Research; Rüschlikon, Switzerland.

[2]ETH; Zürich, Switzerland.

†Currently with Google, Zurich

‡Currently with Axelera AI, Zurich

*Corresponding author. Email: stw@zurich.ibm.com

**Visual oddity task was conceived as a universal ethnic-independent analytic intelligence test for humans. Advancements in artificial intelligence led to important breakthroughs, yet competing with humans on such analytic intelligence tasks remains challenging and typically resorts to non-biologically-plausible architectures. We present a biologically realistic system that receives inputs from synthetic eye movements – saccades, and processes them with neurons incorporating dynamics of neocortical neurons. We introduce a procedurally generated visual oddity dataset to train an architecture extending conventional relational networks and our proposed system. Both approaches surpass the human accuracy, and we uncover that both share the same essential underlying mechanism of reasoning. Finally, we show that the biologically inspired network achieves superior accuracy, learns faster and requires fewer parameters than the conventional network.**

A long-term goal of artificial intelligence (AI) research is to devise cognitive systems that exhibit the capability of abstract reasoning, thus far deemed a distinctive characteristic of human intelligence. In recent years, as deep learning has become more prominent, machine learning models have outperformed humans in several fields, including image classification and playing strategic games [1], [2]. However, competing with humans in tasks where abstract or relational reasoning capabilities are required, is still challenging. Interest in testing human-like reasoning with neural networks has recently grown as more demanding tasks are being considered and large datasets are being created [3]–[7].

In an attempt to tackle these tasks, several neural network architectures have been proposed. A network architecture called Relation Network (RN) was proposed in [8] to achieve state-of-the-art results on the CLEVR dataset [4], a visual reasoning task with textual questions. In [8] it was argued that RNs have a structure suitable for relational reasoning, as they learn to infer an existing relation between objects. The objects, or vector embeddings, are produced by a convolutional neural network (CNN) from an image. All pairs of objects are considered along with an embedding of the question in the task. The model is tailored towards a visual question answering task with a single image. However, an RN-based architecture appears to be useful across all kinds of reasoning tasks. A network called Wild Relation Network (WReN) was proposed in [3] to solve Raven's Progressive Matrices (RPMs), which involve selecting the correct pattern from eight candidate responses to complete a matrix with eight context patterns. This task is difficult for powerful deep networks, including CNNs and recurrent networks [3]. WReN surpasses their accuracy by applying RN modules to infer relations between the elements of matrices. For the solution of problems that require multiple steps of relational reasoning, for example Sudoku riddles, an advanced recurrent RN module has been developed in [9]. A theoretical framework to characterize what tasks a neural network can reason about was developed in [10]. Building on the observation that reasoning processes resemble algorithms, it explains the reasoning capabilities through comparisons of network's computations to reasoning algorithms.

Another important class of tasks requiring analytic intelligence is the so-called visual oddity, first introduced in [11] as part of a neuroscientific experiment to test knowledge of conceptual principles of geometry. This specific visual oddity task is defined on geometrical objects and consists of 45 distinct riddles designed to test basic concepts of geometry such as points, lines, parallelism, and symmetry. Each riddle contains six frames, five of which include a geometrical concept being tested. One of the frames violates the geometrical concept and is called the oddity. The goal is to classify which one of the six frames is the oddity. The task was originally tested on two different groups of people: Americans and Munduruku, an Amazonian indigenous group. Individual task accuracies of both groups were recorded for children and adults. The results showed no significant difference in accuracies between the American and the Munduruku children. This suggests that success in the visual oddity task does not prominently depend on education or experience, thus qualifying the task as an analytic intelligence test. Moreover, it renders the visual oddity task as an appealing abstract reasoning problem to be tackled by AI systems.

Here we take a different perspective on investigating AI systems that can be capable of solving analytic intelligence tasks. Specifically, we take substantial inspiration from biology in several aspects, including neural dynamics, neural architecture and input image processing. Initially, to provide an RN-based baseline, we develop an Oddity Relation Network (OReN) with a task-specific architecture suitable for the solution of the visual oddity task. Then, inspired by biology, we develop a new approach for solving analytic intelligence tasks. To that end, we first focus on spiking neural networks (SNNs) that represent a very efficient class of biologically inspired neural networks [12], [13], passing information through sequences of spikes. Because of the rich dynamics governing the spiking neurons, SNNs are well suited to tackling problems that involve the temporal evolution of the input information fed into the network [14]. Second, we demonstrate that SNNs are also inherently capable of relational reasoning when coupled with a biologically realistic approach to input image processing. In particular, we mimic the saccadic movements made by the human eye during inspection of different images while solving analytic tasks [15]. The SNN dynamics [16] and network architecture are modelled by adopting, as basic building block, the spiking neural unit (SNU) [17] and its soft version (sSNU). This approach is then compared with the OReN for solving the visual oddity task. It turns out that, although the accuracy of neural networks for solving the riddles cannot be directly comparable to that of humans, both extensively trained OReN and SNNs vastly outperform human performance.

### The visual oddity test

The original neuroscientific experimental work [11] refers to a single sample per riddle, consisting of six images. The images were chosen so that their features, which are not tied to the conceptual relation that must be identified to solve the riddle, are not relevant for the solution. For example, these features could be the size and orientation of a segment in images where the middle of segment is the desired concept. The results in [11] indicate that humans did not perform equally well with the various concepts being tested. They performed well with the core concepts of topology, Euclidean geometry, and basic geometrical figures. Recognizing the oddity within geometrical transformations turned out to be significantly more challenging, thus suggesting that such transformations may represent inherently more difficult concepts.

An approach to solve the visual oddity task by resorting to a computational model was proposed in [18], where the authors used the frames from the original work [11] to first generate representations based on glyphs, while separately considering properties of edges, shapes, lines, points, etc. The model then adopts a structure-mapping engine to find the commonality across the frames. This is obtained by relying on analogical generalization to build up a representation of the common features in the images of a riddle. Individual images are compared to the generalization, and the odd image is singled out as the one that exhibits the lowest similarity. In contrast, we put an emphasis on working directly with the raw image data without any additional ad hoc information. In [3], a dataset comprising so-called procedurally generated matrices was developed to solve RPMs. The procedure consists of creating an abstract structure for each matrix by first sampling triples from three primitive sets (relation types, object types, and attribute types) that represent the challenge posed by the matrix. For example, such a triple could consist of progression, shape and number. From the sampled triple, attributes are also sampled from the primitive sets for each type in the triple. However, determining a similar structure for the visual oddity task is infeasible, because of the variability of the frames within a single riddle and across the 45 different riddles. Thus, in order to use supervised learning methods for the visual oddity task a large sample dataset is critical. To that end, we created a dataset where different samples of each riddle with a specific underlying geometrical concept are procedurally generated. Twelve sample generated riddles representing different core concepts are illustrated in Fig. 1 along with the corresponding averaged percentage of correct answers for human participants as reported in [11]. The generation procedure is explained in Methods.

As mentioned earlier, the visual oddity task tests the capability of making analogies by relying on geometric concepts such as points, lines and angles. Therefore, the detection of similarities is key to solving the riddles. For humans, abstract reasoning is crucial to find a correct answer, as no two images are the same and there are several ways to detect an odd image. In fact, to succeed, the participants must infer the geometrical relationship between the array of images. Therefore, the challenge for neural networks lies in whether they can infer the abstract geometrical relationships between the images. Training different neural network architectures to solve the visual oddity task provides an insight into their relation-forming capabilities.

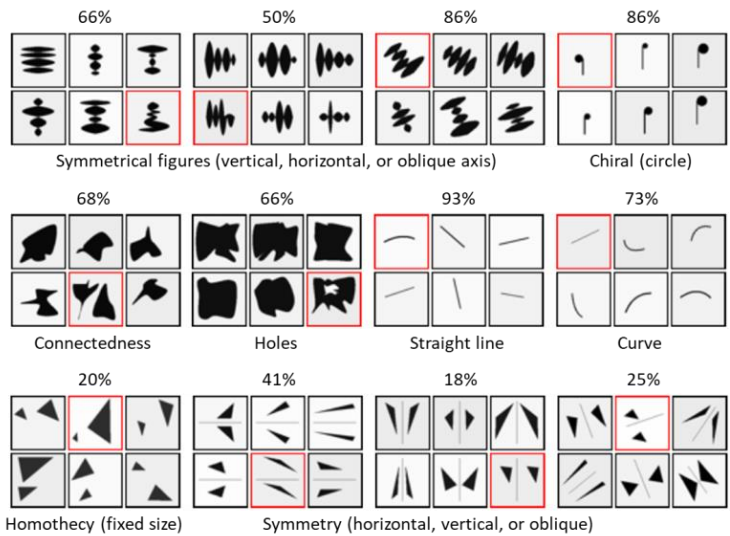

**Figure 1. Generated visual oddity riddles.** A set of twelve visual oddity riddles representing different core concepts and the corresponding averaged percentage of correct answers for human participants.

**Oddity Relation Network (OReN)**

RNs were proposed in [8] as a general solution to relational reasoning in neural networks that extract information about a relation that may be formed by all pairs of objects in certain contexts. They exhibit three main characteristics, which make them well suited for solving relational reasoning tasks. First, these networks are capable of learning to infer the existence of object relations, as all pairs of objects and their potential relations are considered. Second, they compute each relation by a single function, typically a multi-layer perceptron (MLP), which leads to high data efficiency. The MLP operates on a batch of object pairs, where each sample of the batch is drawn from the same set. Third, they operate on a set of objects for which a specific order is not required, thus ensuring that the RN's output contains information that reliably represents the relations within the set.

As mentioned in the introduction, the WReN is an RN that was specifically proposed to solve RPMs [3]. Consequently, it is particularly appropriate to address reasoning tasks that require the detection of similarity between images. The model computes pair-wise relations between context panels and response panels in an RPM dataset to infer the relations between the eight matrix elements and the eight candidate answers. The information pertaining to context-context relations and context-multiple-choice relations is then integrated to provide a score for the selection of the answer. For the visual oddity task, there are neither context panels nor choice panels, as all frames can be classified as the oddity. Therefore, we modify the WReN to obtain the OReN and investigate its capability to solve all 45 riddles of the visual oddity task. In the OReN, first the pair-wise relations between each panel and the remaining five panels are formed, then the resulting information is integrated to obtain a score for the identification of the oddity.

For the experiments, we use datasets generated as described in Methods. Each sample in a dataset comprises six frames and a label, which indicates the index of the oddity. Each frame is first input into a 5-layer CNN, referred to as the vision model. For every frame $k$, $k \in [1, 2, ..., 6]$, the vision model outputs a frame embedding $\eta_k$, represented as a vector with $D$ dimensions. For each frame embedding $\eta_k$, pairs are generated by ordered concatenation with all six frame embeddings. A total of 36 pairs are thus generated. Each pair is input to a function, $g_\theta$, parameterized by a neural network $\theta$. The six outputs of $g_\theta$ corresponding to frame $k$ are summed up for each frame. The summed output of each frame is input to a second function, $f_\varphi$, also parameterized by a neural network $\varphi$, to calculate the final score. The score for each frame $k$ is then calculated as

$$q_k = f_\varphi \left( \sum_{i=1}^{6} g_\theta \left( \eta_k, \eta_i \right) \right).$$

A softmax function is finally applied across all scores to determine the probability of each frame being the oddity. A generic architecture for the OReN is illustrated in Fig. 2, with details described in Methods. This figure also depicts the generation of the vector embeddings from the images of a riddle by the vision model.

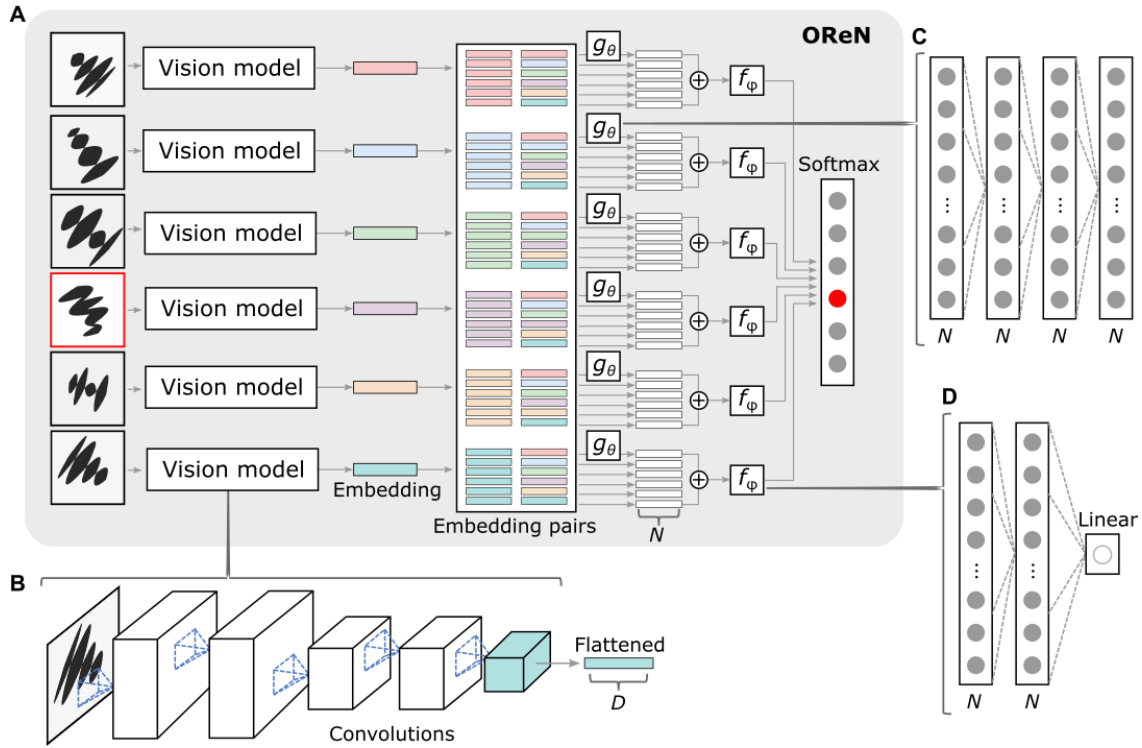

**Figure 2. OReN architecture.** (**A**) Generic architecture showing the vision model, the formation of embeddings and embedding pairs, as well the relation and decision stages. (**B**) Realization of the vision model by a CNN. (**C**) Realization of the function $g_\theta$ through a four-layer MLP with $N$ rectified linear neurons per layer. (**D**) Realization of the function $f_\varphi$ with a three-layer MLP with two $N$-sized rectified linear layers and a single linear output neuron yielding the score.

**Saccadic neural network**

The human approach to analytic reasoning differs from the operation of relational networks, such as OReN, starting with the manner the input stimuli are delivered from the eyes. Solving analytic reasoning tasks, where the inspection of various images is required, involves a series of repeated back-and-forth rapid eye movements, called saccades. The saccades allow the brain to compare all candidate images over a certain period and reach a conclusion, as observed in human subjects solving the RPM task [15]. Moreover, the operation of the biological neurons in the brain is characterized by spike-based communication and rich temporal neural dynamics that is often abstracted into a so-called leaky integrate and fire (LIF) neuron model, used widely in SNNs. Therefore, it is appealing to explore analytic reasoning for the oddity task taking direct inspiration from a more biologically realistic approach.

To this end, we synthesize a series of eye saccades over the candidate frames and input them into a temporal saccadic neural network model, as illustrated in Fig. 3A. To ensure a balanced distribution of the vision input stimuli, i.e., that each candidate frame is observed the same number of times, we construct the input as a series of six random permutations of all six candidate frames, resulting in 36 time steps simulating a series of saccades. At each time step $t$, the saccadic network outputs its belief that the currently presented frame is the oddity: $p(oddity|t)$. During testing, the initial $S_I=18$ saccades are used to initialize the temporal dynamics of the reasoning neurons, whereas the probabilities obtained during the $S_E=18$ evaluation saccades are integrated in order to provide the final decision.

The architecture of the saccadic network model is biologically inspired in several aspects. The vision model is a CNN that is reminiscent of the receptive fields in the visual cortex [19]. For a fair comparison with OReN, the same CNN structure is used in the saccadic network. Then, the recurrent neural units implement the temporal dynamics found in neocortical neurons. Specifically, as illustrated in Fig. 3B, the LIF abstraction of the biological neurons is simulated using a deep learning recurrent unit following the SNU approach [17]. The state equations for a layer of SNUs are

$$s_t = g(Wx_t + Hy_{t-1} + \lambda \odot s_{t-1} \odot (1 - y_{t-1}))$$
$$y_t = h(s_t + b),$$

where $x_t$ denotes the input at time $t$, $s_t$ indicates the internal state modelling the membrane potential $V_m$, $y_t$ denotes the output, and $g$ and $h$ represent the input and output activation functions, respectively. Furthermore, $W$ and $H$ represent input and recurrent synaptic weight matrices, $\lambda$ denotes a leak parameter, $b$ is a bias that determines the firing threshold, and $\odot$ denotes point-wise multiplication. For the saccadic network, we employ three SNU layers, with $N$ units each, followed by a single sigmoid readout neuron, that provides the belief value at its output. This formulation, which is relying on a deep learning approach, enables to operate in the spike-based SNN mode with SNUs assuming $h=\Theta$, the Heaviside function, as well as in the standard artificial neural network mode with soft SNUs assuming $h=\sigma$, the sigmoid function, or even to use the popular LSTM units. This allows us also to compare the performance of the biologically inspired neural units with that of common deep learning recurrent units.

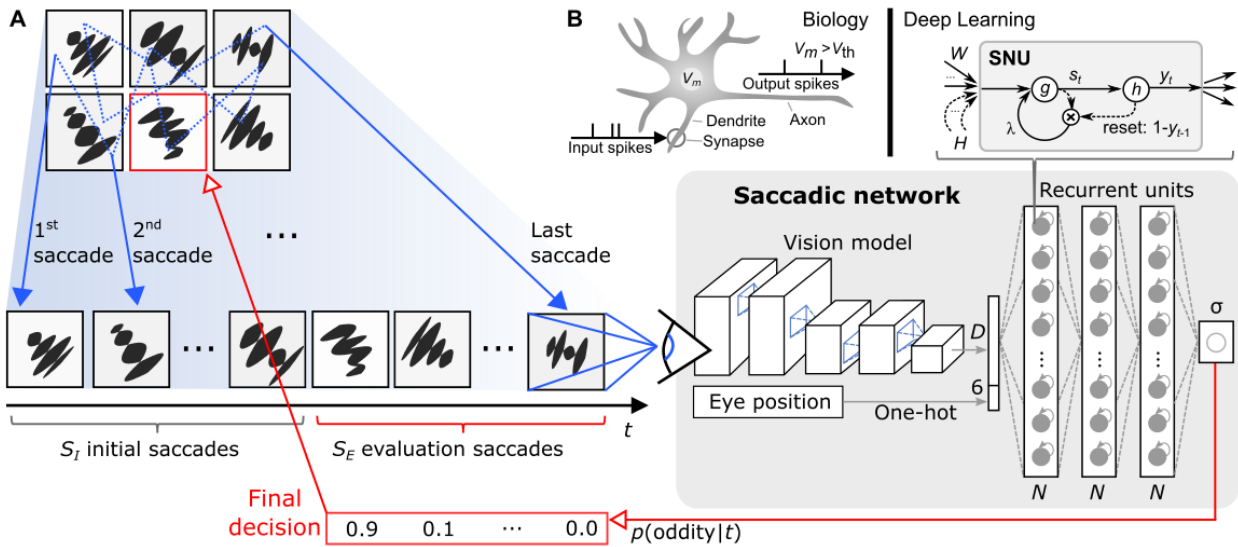

**Figure 3. Visual analytic reasoning approach with a saccadic network.** (**A**) The network receives image embeddings and eye-position inputs from a series of simulated saccadic eye movements and performs reasoning using the internal temporal dynamics. After $S_I$ initial saccades, the final decision is determined from the network outputs over $S_E$ evaluation saccades. (**B**) The neural dynamics of biological neurons are incorporated into a deep network through SNUs, where each SNU represents an abstraction of the LIF model.

The key feature of the stateful neurons considered here consists in the processing of the current input within the context of the neuronal state that reflects the information from the past inputs. In other words, both the comparison of embeddings and the accumulation of information is performed over time, rather than over space as is done in RNs. Importantly, we show that the obtained membrane potential integration, in conjunction with a biologically realistic saccadic input stream, enables analytic reasoning. Common stateless deep networks struggle to solve analytic reasoning tasks. Our initial attempts to solve the oddity task with fully connected networks or even ResNet architectures led to ~17% accuracy, which corresponds to guessing by chance. The essential functionalities introduced in the RN architecture, which enable analytic reasoning in such networks, are the addition of the processed representations and the reuse of the $g_\theta$ and $f_\varphi$ modules for all the frames. Apparently, the first functionality is inherently present in neural recurrent units, such as SNUs, where an input $x_t$, transformed by the weights $W$ is added to the previously processed inputs represented in the state $s_{t-1}$. Furthermore, the idea of RN to process a series of inputs by replicating the same parameters in space, see Fig. 2, is naturally realized by a series of saccadic movements, where each input is processed by the same units, thus reusing the same synaptic weights $W$. The two biologically realistic aspects, namely the stateful operation in combination with a saccadic input sequence, provide the means for analytic reasoning.

**Can biologically inspired neural networks compete with relation networks?**

We compared the accuracy of OReN and saccadic networks in two setups. First, in a separate training setup, in which networks were trained and evaluated for each task separately. Second, in a joint training setup, in which a single network was trained and evaluated on all tasks. In both setups, we varied the layer size $N$ following a geometric progression, from 16 to 256 and from 64 to 4096, respectively. For a saccadic network we considered four types of units in the reasoning part: SNU-based spiking LIF neurons (SNN), soft SNUs (sSNU), soft SNUs with layer-wise recurrency (sSNU-R), and standard LSTMs. Details are described in Methods. The results for the separate setup are presented in Fig.4A. OReNs achieved an impressive averaged accuracy of 98.3% for $N$=128 (~1.0M model parameters). This demonstrates that the RN-based approach enables oddity tasks to be effectively solved. Saccadic networks with LSTM units performed slightly worse with 98.2%, while requiring many more parameters, ~4.7M for $N$=256, for legibility reasons not shown within the plot boundary – see Supplementary Note 3 for a table with comprehensive results. Remarkably, the saccadic networks operating with much simpler biologically inspired dynamics achieved the best result of ~99% using the fewest parameters: ~0.5M ($N$=128) for SNN, ~0.3M ($N$=64) for sSNU and ~0.2M ($N$=32) for sSNU-R. Fig. 4B illustrates the results for the joint training setup, which is more challenging as the network needs to seamlessly switch between recognizing 45 different kinds of oddities. In consequence, larger models were required to learn well, and the final accuracy levels dropped by ~2%. OReN in the best performing configuration of $N$=1024 (~12M) obtained 95.2%, which is a very good result in absolute terms. However, saccadic networks performed better, reaching a top accuracy of 97.0% for sSNU-R with $N$=2048 (~28M). SNN and sSNU achieved 95.3% and 96.3%, respectively. In comparison, LSTM performed quite well for the joint setup, achieving 96.4% for $N$=512 (~12M).

The accuracy achieved by OReNs and saccadic networks is very high, especially when contrasted with 66.8% average accuracy for humans in the original experiment [11] and with 86.7% accuracy for the computational model [18], whose accuracy deteriorated for tasks considered difficult for humans. Fig. 4C depicts the tasks sorted by increasing difficulty level for humans along with the accuracy for the most biologically plausible of our models: a saccadic network with spiking neurons ($N$=128). Although for some more difficult tasks, e.g. task 31 or 33, the final accuracy indeed dropped, a very weak correlation was observed in terms of the final accuracy-dependence on the task difficulty index. Furthermore, these accuracy values are not directly comparable to human accuracy, because of a markedly different setup of the experiments. Humans solve the oddity tasks through generalization of analytic reasoning principles acquired in various contexts throughout their lives, whereas in our setup the networks were trained to discover these principles from examples directly reflecting the oddity tasks. Nevertheless, the accuracy of the SNN is above the measured human-level accuracy for all tasks. Fig. 4D illustrates how many training epochs were required to reach the human accuracy by the SNN saccadic model. On average it required ~3.5 epochs, corresponding to 8,932 examples. For comparison, the best OReN model with $N$=128 required ~4.1 epochs, corresponding to 10,524 examples. In this context, human performance remains quite impressive.

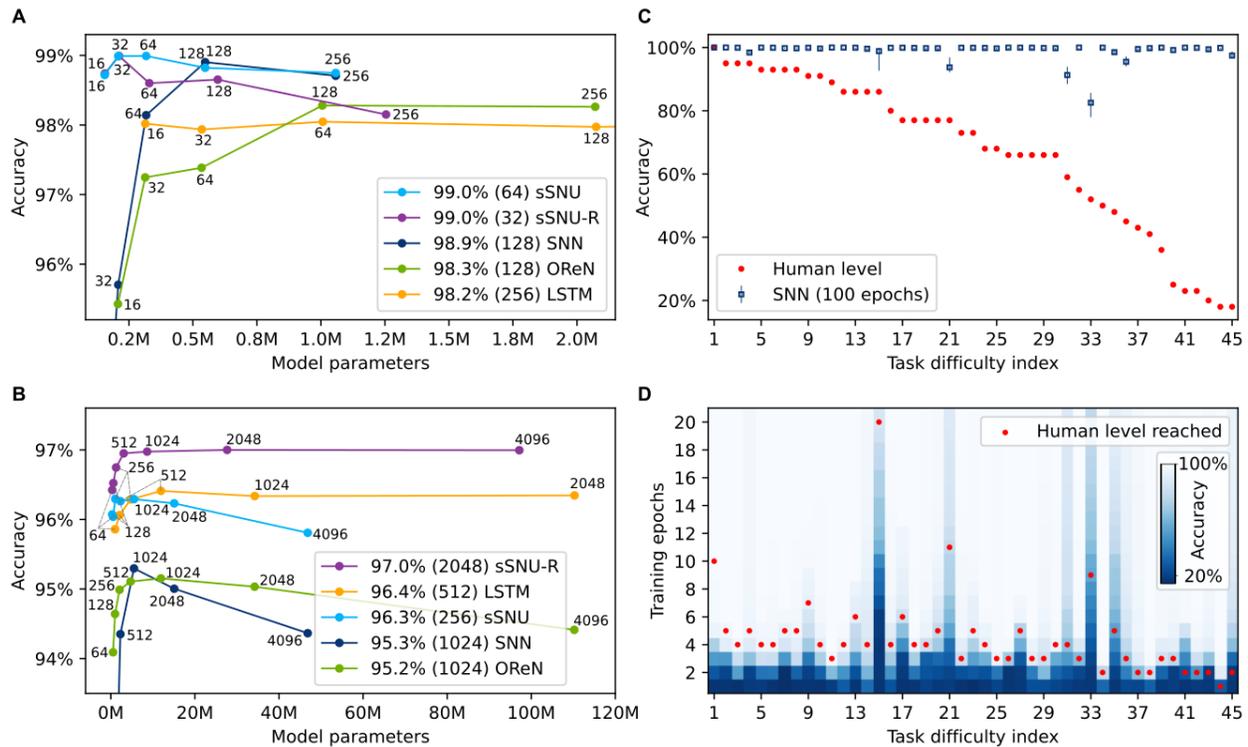

**Figure 4. Accuracy comparison.** (**A**, **B**) Average test accuracy for different model sizes ($N$ value reported next to each point) for separate (A) and joint (B) training setup. Legend labels report the best averaged accuracy and the corresponding $N$ value for each series. See Supplementary Note 3 for detailed values. (**C**) SNN ($N$=128) separate training accuracy with min/max error bars plotted along with the human-level accuracy, for tasks sorted by decreasing difficulty level for humans, and (**D**) test accuracy evolution over the first 20 training epochs with marked epochs when reaching human-level accuracy.

Although OReNs and saccadic networks seemingly operate based on different principles, the discussion in the previous section unveiled similarities between the two approaches. We conjectured that an OReN performs comparisons in space, whereas the saccadic network does so over time. We validate these statements by visualizing the internal representation during inference of trained OReNs and saccadic networks with SNNs, with $N=32$ for both. The activity maps calculated in an OReN through $g_\theta$ and $f_\varphi$ are depicted in Fig. 5A. The second activity map is visually distinct from others, and after the application of the scoring function $f_\varphi$, the OReN correctly identifies that the second input image is the oddity. Next, the same set of images is presented to the saccadic network through a series of saccades. The state values of Layer 3 are visualized in Fig. 5B. Gradual and abrupt decrease of $V_m$ over time (the time axis is vertical) corresponds to membrane potential leakage with parameter $\lambda$ and resetting after spikes, respectively. In this form, it is difficult to compare both models. However, as seen in Fig. 5C, when membrane potentials are sorted as if the input saccades would follow the order of the first items from the OReN embedding pairs, the membrane potentials of Layer 3 neurons become visually distinctive for the saccades comprising the oddity, similarly to the activity maps in the OReN. Both models correctly identify the second image as the oddity based on the same principle of comparisons. In the OReN the comparisons are performed in space, with each pair of input images explicitly presented at different inputs of the network at the same time. In the saccadic network, the temporal dynamics implicitly stores the context and it is possible to operate with only one input image observed at a time. This is similar to how the brain analyzes the sensory inputs from saccades over time and compares information to the past context to perform analytic reasoning.

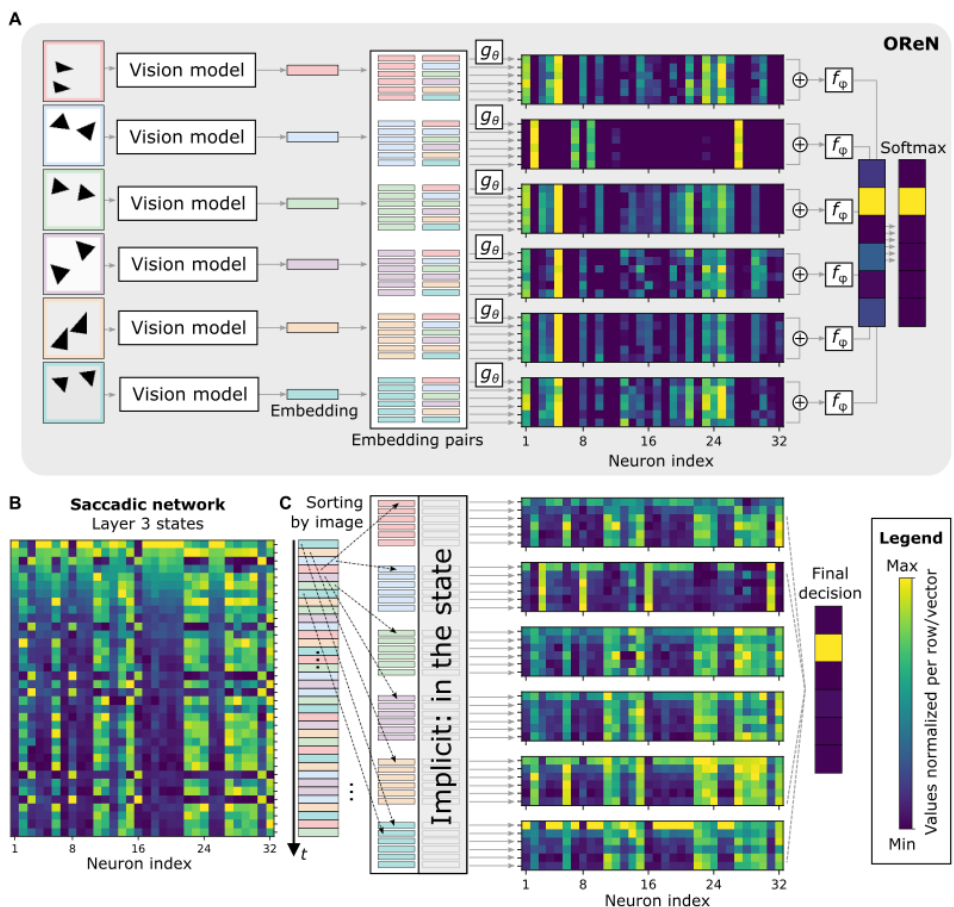

**Figure 5. Activity comparison of an OReN and a saccadic network using spiking neurons**. $N=32$. (**A**) OReN identifies the oddity through calculation of $g_\theta$ for pairs of input embeddings, followed by aggregation and scoring through $f_\varphi$. The embedding pairs comprising the oddity as the first element lead to a visually distinctive $g_\theta$ activation map for the oddity. (**B**) The saccadic network receives a temporal stream of inputs that lead to the evolution of the internal states, $V_m$, visualized for Layer 3. (**C**) Membrane potentials sorted by the order of the first OReN embedding pair item exhibit similar visual distinction for the oddity. Visualized values were normalized row-wise, i.e. over 32 neurons.

**Summary**

We investigated the task of visual oddity, which requires a high level of visual analytic intelligence, from the perspective of both RNs and biologically inspired saccadic architectures. We first generated a large dataset of visual oddity riddles that represent the core concepts to be tested and allow thorough network comparisons. Then we explored two avenues, one where the RN approach was extended to give origin to an OReN capable of tackling visual oddity tasks, the other where a novel biologically inspired architecture processes a stream of input images to determine the oddity. Both avenues lead to networks that are characterized by the capability of establishing relations between the images of a riddle, and outperform previously proposed computational models, as well as previously reported human performance, the latter being achieved though in the context of lifetime experience. Furthermore, we validated the conjecture that biologically inspired saccadic networks evaluate the image relations sequentially over time, whereas the RNs operate in parallel over space. Our findings indicate that saccadic network models achieve better performance than RNs and require a significantly lower number of model parameters.

**References:**

[1]   O. Russakovsky *et al.*, "Imagenet large scale visual recognition challenge," *Int. J. Comput. Vis.*, vol. 115, no. 3, pp. 211–252, 2015.

[2]   D. Silver *et al.*, "Mastering the game of Go with deep neural networks and tree search," *Nature*, vol. 529, no. 7587, pp. 484–489, Jan. 2016, doi: 10.1038/nature16961.

[3]   D. Barrett, F. Hill, A. Santoro, A. Morcos, and T. Lillicrap, "Measuring abstract reasoning in neural networks," *Proc. ICML*, pp. 511–520, 2018.

[4]   J. Johnson, B. Hariharan, L. Van Der Maaten, L. Fei-Fei, C. Lawrence Zitnick, and R. Girshick, "Clevr: A diagnostic dataset for compositional language and elementary visual reasoning," *Proc. CVPR*, pp. 2901–2910, 2017.

[5]   H. J. Levesque, E. Davis, and L. Morgenstern, "The Winograd Schema Challenge," in *Proceedings of the Thirteenth International Conference on Principles of Knowledge Representation and Reasoning*, Rome, Italy, 2012, pp. 552–561.

[6]   N. Mostafazadeh *et al.*, "A corpus and cloze evaluation for deeper understanding of commonsense stories," in *Proceedings of the 2016 Conference of the North American Chapter of the Association for Computational Linguistics: Human Language Technologies*, San Diego, California, 2016, pp. 839–849. doi: 10.18653/v1/N16-1098.

[7]   J. Weston *et al.*, "Towards AI-complete question answering: A set of prerequisite toy tasks," *Proc. ICLR*, 2016.

[8]   A. Santoro *et al.*, "A simple neural network module for relational reasoning," *Adv. Neural Inf. Process. Syst.*, vol. 30, 2017.

[9]   R. Palm, U. Paquet, and O. Winther, "Recurrent relational networks," *Proc. NeurIPS*, vol. 31, 2018.

[10]  K. Xu, J. Li, M. Zhang, S. S. Du, K. Kawarabayashi, and S. Jegelka, "What can neural networks reason about?," *Proc. ICLR*, 2020.

[11] S. Dehaene, V. Izard, P. Pica, and E. Spelke, "Core Knowledge of Geometry in an Amazonian Indigene Group," *Science*, vol. 311, no. 5759, pp. 381–384, Jan. 2006, doi: 10.1126/science.1121739.

[12] S. Ghosh-Dastidar and H. Adeli, "Third generation neural networks: Spiking neural networks," in *Advances in Computational Intelligence*, Springer, 2009, pp. 167–178.

[13] F. Ponulak and A. Kasinski, "Introduction to spiking neural networks: Information processing, learning and applications.," *Acta Neurobiol. Exp. (Warsz.)*, vol. 71, no. 4, pp. 409–433, 2011.

[14] A. V. Herz, T. Gollisch, C. K. Machens, and D. Jaeger, "Modeling single-neuron dynamics and computations: a balance of detail and abstraction," *Science*, vol. 314, no. 5796, pp. 80–85, 2006.

[15] T. R. Hayes, A. A. Petrov, and P. B. Sederberg, "A novel method for analyzing sequential eye movements reveals strategic influence on Raven's Advanced Progressive Matrices," *J. Vis.*, vol. 11, no. 10, pp. 10–10, Sep. 2011, doi: 10.1167/11.10.10.

[16] W. Gerstner, W. M. Kistler, R. Naud, and L. Paninski, *Neuronal dynamics: From single neurons to networks and models of cognition*. Cambridge University Press, 2014.

[17] S. Woźniak, A. Pantazi, T. Bohnstingl, and E. Eleftheriou, "Deep learning incorporating biologically inspired neural dynamics and in-memory computing," *Nat. Mach. Intell.*, vol. 2, no. 6, pp. 325–336, Jun. 2020, doi: 10.1038/s42256-020-0187-0.

[18] A. Lovett, K. Lockwood, and K. Forbus, "A computational model of the visual oddity task," in *Proceedings of the Annual Meeting of the Cognitive Science Society*, 2008, vol. 30, no. 30.

[19] K. Fukushima, "Neocognitron: A self-organizing neural network model for a mechanism of pattern recognition unaffected by shift in position," *Biol. Cybern.*, vol. 36, no. 4, pp. 193–202, Apr. 1980, doi: 10.1007/BF00344251.

**Methods:**

**Generation of tasks.** To use standard supervised learning methods to solve the visual oddity task we need a dataset for each of the 45 different tasks. As the original paper [11] only contains one sample per task, we created a dataset where each task is procedurally generated. Below we describe the task generation procedure. Each of the 45 different tasks has individual attributes. For the dataset generation process, we consider the attributes of each task in a procedural way. In order to ensure as much variability as possible in the dataset, we include all the possible attributes as variables that take randomized values. Each sample in the dataset is a combination of six 100×100 frames and a label, which indicates the index of the oddity in the array. Every frame we generate is an 8-bit grayscale figure. Each pixel has 256 possible values ranging from 0 (black) to 255 (white). Each frame has a background grayscale value ranging from 235 to 255. For each frame, we additionally randomly sample a grayscale value between 0 and 61 to be used for all surfaces, edges and points. The generated frames are compared with each other to ensure that they are all pixel-wise unique. We generate 45 small datasets of size 3,840 separately for each specific oddity type, and one large dataset of size 108,000 comprising all oddity types. Each dataset is split into training, validation and test set, with sample count following a ratio of 4:1:1, respectively.

When generating a dataset for a single task, the task id (ranging from 1 to 45) is fixed, so it will be the only task for which frames are generated. In contrast, when generating a dataset containing multiple tasks, the task id is sampled randomly from the set of tasks for each sample in the dataset. Each task has its own generator that first generates five frames that contain the common conceptual geometrical property of the task, called non-oddities. Afterward, the task generator generates one frame that does not contain the property, i.e., the oddity. The six frames are then randomly shuffled and the index of the oddity is returned as the label. All sizes are determined in a percentage of the maximum frame size to allow the generation of various frame sizes. An example of each generator output can be seen in Supplementary Figure 1, where the oddity of each task is also highlighted.

All 45 tasks are divided into eight categories as defined in [11], where each category corresponds to one of the considered geometrical challenges. The details of each category along with the variables used in each task are listed in Supplementary Note 1. Some variables are used for every task in a category and some are used for several categories, as indicated in Supplementary Tables 1-8.

**Detailed architecture of OReN.** The OReN architecture, depicted in Fig. 2A, was implemented in TensorFlow 1. It comprises a CNN vision model that is applied six times to produce 3200-dimensional embeddings of the six candidate frames. The embeddings are paired into 6400-dimensional vectors that are processed by the MLP $g$. The resulting vectors are summed and scored by a second MLP $f$. The oddity is determined by detecting which neuron yields the highest value from the final softmax layer.

The CNN vision model, depicted in Fig. 2B, receives an input image rescaled to 80×80 and processes it using five 32-channel convolutional layers with 5×5 kernels and no border padding. Each convolution is followed by a batch normalization. The first, third and the fifth convolution layers are followed by a dropout operation with 0.3 dropout rate. The second and the fourth convolution layers are followed by 2×2 max pooling with horizontal and vertical strides equal to two, and zero padding at the borders. The output of the final convolutional layer is flattened into a 3200-dimensional embedding vector. A detailed visualization of the vision model is included in Supplementary Note 2.

The $g$ function is instantiated for a series of 6400-dimensional vectors comprising concatenated pairs of embeddings, as depicted in Fig. 2C. In each case, the vector is processed by a four-layer fully connected network with $N$ rectified linear units per layer. Each layer is followed by a dropout operation with 0.3 dropout rate. The output is an $N$-dimensional vector. For each embedding pair group, six such vectors are added and form an input to the $f$ function, as illustrated in detail in Fig. 2D. This input is processed by two fully connected layers with $N$ rectified linear units per layer, followed by a single linear neuron that produces the final score for the considered group of the embedding pairs. The six scalar scores from the six instantiations of $f$ are processed by a softmax layer that provides the final probability distribution, where the neuron yielding the highest value indicates the index of the oddity. A categorical cross-entropy loss between the output probabilities and the ground-truth indices of the oddities is minimized during training using the Adam optimizer.

**Detailed architecture of saccadic network.** The saccadic network architecture, depicted in Fig. 3B, was implemented in TensorFlow 1 and comprises the same CNN vision model as the OReN network. For each saccade, the vision model computes a 3200-dimensional embedding that is concatenated with the current eye position encoded as a one-hot vector (all zeros except the position corresponding to the index of the currently examined panel). The concatenated vectors are processed by three consecutive fully-connected layers with $N$ recurrent units.

The recurrent units are either LSTMs or SNUs. The default TensorFlow 1 configurations of activation functions, parameter initializers and other settings are used unless stated otherwise. The specific hyperparameters for SNUs are the input activation function $g$ that is set to identity function (no input activation) and leak parameter $l$ that is set to 0.8. SNUs in SNN configuration have the activation function $h$ set to the step function and in sSNU configuration it is set to the sigmoid function. SNUs with "-R" suffix include recurrent connections matrix $H$, that is skipped otherwise. The bias $b$ is initialized to -1.0.

The outputs from the recurrent units are processed by a single stateless sigmoidal output neuron that provides the output value for the current saccade. A series of saccades leads to a sequence of sigmoidal outputs, for which the error is minimized using a binary cross-entropy loss and Adam optimizer. The binary cross-entropy loss is masked, so that it considers only the relevant saccades. The outputs for the first two saccades are masked, as the appearance of the third distinct frame is the earliest possible moment when it becomes possible to conjecture which frame is the oddity.